\documentclass[letterpaper, 10 pt, conference]{ieeeconf}  

\IEEEoverridecommandlockouts                              

\usepackage{graphicx}
\usepackage{hyperref}
\usepackage{subcaption}
\usepackage{url}
\usepackage{tikz}
\usepackage{fancyhdr}

\title{\LARGE \bf
A Standing Support Mobility Robot for Enhancing Independence in Elderly Daily Living
}

\author{Ricardo J. Manríquez-Cisterna$^{1}$, Ankit A. Ravankar$^{1}$, Jose V. Salazar Luces$^{1}$,\\ Takuro Hatsukari$^{2}$, and Yasuhisa Hirata$^{1}$
\thanks{$^{1}$Graduate School of Engineering, Department of Robotics, Tohoku University, Sendai 980-8579, Japan.
        {\tt\small Emails: \{r.manriquez, ankit, j.salazar, hirata\}@srd.mech.tohoku.ac.jp}}%
\thanks{$^{2}$ Paramount Bed Co. Ltd.
        Higashiuna, Koto, Tokyo 136-8670, Japan
        {\tt\small t.hatsukari@paramount.co.jp}}%
}

\begin{document}

\renewcommand{\baselinestretch}{0.99}
\maketitle

\fancypagestyle{firstpage}{%
  \fancyhf{} 
  \renewcommand{\headrulewidth}{0pt} 
  \renewcommand{\footrulewidth}{0pt} 
  \fancyfoot[C]{\footnotesize
    This work has been submitted to the IEEE for possible publication.
    Copyright may be transferred without notice, after which this version
    may no longer be accessible. This work has been accepted for the 34th IEEE International Conference on Robot and Human Interactive Communication, IEEE RO-MAN 2025.}%
}
\thispagestyle{firstpage}
\pagestyle{empty}

\begin{abstract}
This paper presents a standing support mobility robot ``Moby" developed to enhance independence and safety for elderly individuals during daily activities such as toilet transfers. Unlike conventional seated mobility aids, the robot maintains users in an upright posture, reducing physical strain, supporting natural social interaction at eye level, and fostering a greater sense of self-efficacy.  Moby offers a novel alternative by functioning both passively and with mobility support, enabling users to perform daily tasks more independently. Its main advantages include ease of use, lightweight design, comfort, versatility, and effective sit-to-stand assistance. The robot leverages the Robot Operating System (ROS) for seamless control, featuring manual and autonomous operation modes. A custom control system enables safe and intuitive interaction, while the integration with NAV2 and LiDAR allows for robust navigation capabilities. This paper reviews existing mobility solutions and compares them to Moby, details the robot’s design, and presents objective and subjective experimental results using the NASA-TLX method and time comparisons to other methods to validate our design criteria and demonstrate the advantages of our contribution.\\
Video: \url{https://bit.ly/moby-robot}
\end{abstract}
\vspace{0.5em}


\section{INTRODUCTION}
As global life expectancy continues to rise, societies around the world are confronting the challenge of supporting an aging population with limited caregiving resources \cite{bloom2010population}. This issue is particularly pronounced in countries like Japan, where nearly one in three individuals will be aged 65 or older by year 2036 \cite{EPRS2020,Chiba2024}. The imbalance between the growing number of older adults and the availability of caregivers has led to a pressing need for technological solutions that support independent living and reduce physical strain on human caregivers \cite{yamazaki2021frailty}.
\begin{figure}[!ht]
    \centering
    \includegraphics[width=0.7\linewidth]{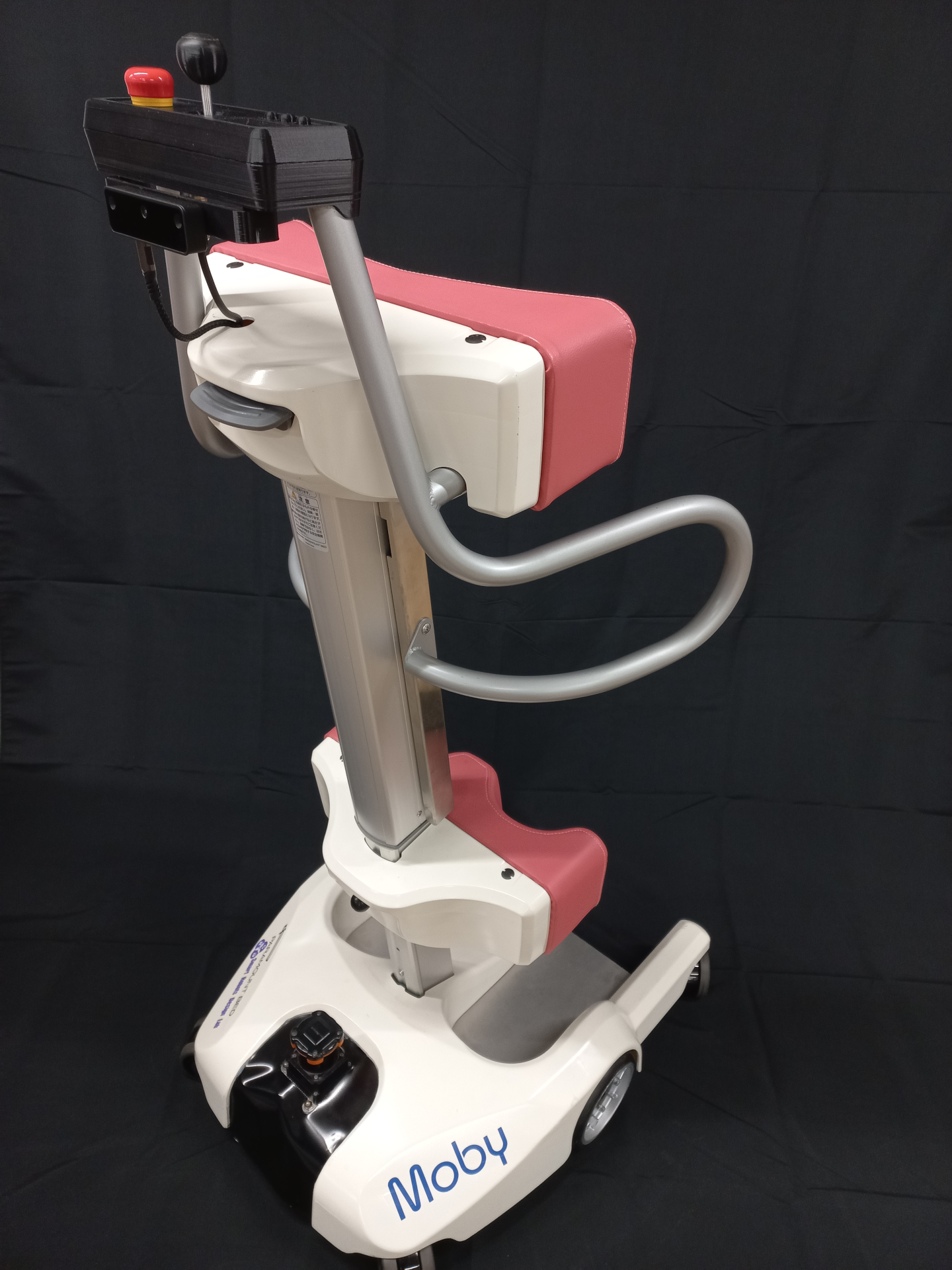}
    \caption{Moby - A standing mobile support robot with emphasis on sit to stand and elderly care}
    \label{fig:figure-moby}
\end{figure}
 Several research has been ongoing on mobility support and assist robots for elderly and rehabilitation in recent years \cite{wang2024socially,ruiz2021improving,terayama2024concept,bemelmans2012socially,Bateni2005mobilityaids}. These mobility devices range from passive support devices,  active support devices or hybrid support devices providing different levels of support based on individual needs and demands. For example, wheelchairs, offer a seated mobility, and are widely used \cite{gowran2022wheelchair}. However, the transition to a wheelchair can often mark a psychological turning point, symbolizing a loss of independence and potentially accelerating physical deconditioning \cite{Sonenblum02112021}. 
 There exists a largely underserved space between these options—users who retain some standing and walking ability, but require assistive support to do so safely and confidently. To fill this gap, we introduce Moby (from mobility), a robotic mobility aid designed to support sit-to-stand transitions, standing posture, and short-distance walking. \autoref{fig:figure-moby} shows the Moby platform. Moby’s design emphasizes user safety, adaptability to different body types, and ease of operation in home and care environments. Unlike passive aids, Moby offers a semi-active platform that can bridge the gap between full independence and total dependence, supporting a more dignified aging experience. The robot is equipped with multipurpose handlebars, continuous height adjustment, and adjustable shin support, allowing it to accommodate users of different body sizes and mobility levels. The ability to customize these features ensures that Moby can be tailored to each individual, supporting the goal of empowering users to maintain their autonomy in daily activities. Furthermore, the robot’s 100 kg payload capacity ensures that it can assist a wide range of users, from those with lower mobility to those requiring a bit more support.

Moby was designed with a clear emphasis on safety, simplicity, and adaptability. The system enables users to perform sit-to-stand transitions while providing real-time physical support and stabilization throughout the process. By preserving an upright posture and supporting short-distance walking, Moby helps maintain not only physical capabilities but also social interaction and engagement—factors that are often compromised in traditional mobility solutions.

The robot is intended for use in a range of environments, including homes, rehabilitation centers, and long-term care facilities, where individualized mobility support is increasingly needed \cite{ravankar2023real}. Its user-centric design allows caregivers to assist more efficiently and reduces the likelihood of injuries during manual transfers.
Our aim is to contribute to the growing field of assistive robotics with a functional, deployable system tailored to the needs of an aging society.

In this paper, we present the design and development of Moby as a functional, deployable assistive robot. We detail its mechanical configuration, control system, and safety features, and evaluate its usability through real-world testing with diverse users. Our results highlight how robotic assistance can enhance mobility while supporting a more dignified and independent lifestyle for older adults and individuals with physical limitations.




\subsection{Related Works}
Traditional mobility aids, such as canes and walkers, offer basic support but often require significant upper-body strength and provide limited stability, particularly for complex transitions \cite{Bateni2005mobilityaids,martins2012assistive,bradley2011geriatric}. Wheelchairs, while providing seated mobility, can restrict upright posture, reduce physical engagement, and may psychologically impact a user's sense of independence \cite{haisma2006physical}.

In recent years, research has increasingly focused on robotic mobility support and assist robots for elderly and rehabilitation purposes. For instance, socially assistive walkers are being developed to provide daily living assistance to older adults \cite{bemelmans2012socially}. Mobile collaborative robots have also been explored to improve standing balance performance \cite{maurice2017human}. Other works investigate adaptive touch walking support robots designed to maximize human physical potential \cite{terayama2024concept}. These systems demonstrate the growing interest in leveraging robotics to support an aging population.

However, there remains a largely underserved space for users who retain some standing and walking ability but require assistance to do so safely and confidently. A key factor in the effectiveness of such systems lies not only in physical support, but also in the preservation of \textit{self-efficacy}—the belief in one’s ability to perform tasks independently \cite{wen2022sense}. Studies have shown that self-efficacy strongly influences well-being and autonomy in elderly individuals \cite{hirata2022cooperation,bandura1997selfefficacy,ravankar2022care}. Maintaining abilities such as standing, walking, and transferring helps reinforce a sense of control and contributes positively to mental and emotional health.
There exists a largely underserved space between these options—users who retain some standing and walking ability, but require assistive support to do so safely and confidently. 

Beyond direct physical assistance, the integration of autonomous capabilities, particularly in robot mapping and navigation, is crucial for seamless deployment in complex environments like homes and care facilities \cite{simpson2008many,ravankar2020safe,levine1999navchair,  beomsoo2021mobile}. Advancements in robust robot mapping and navigation techniques \cite{cadena2016past} enable robots like Moby to move safely and efficiently within unconstrained indoor environments, reducing the need for continuous human supervision.

Few existing devices effectively preserve an upright posture while offering real-time, user-adaptive physical assistance, adjustability, and ease of use in a unified system. Unlike many existing robotic walkers that primarily focus on gait assistance, Moby uniquely emphasizes active sit-to-stand transitions and sustained upright posture during short-distance mobility, particularly for critical daily tasks like toileting, thereby bridging the gap between full independence and total dependence.
\section{Design}

Moby is a semi-passive assistive robot developed to support users with reduced lower-limb strength during posture transitions and short-distance mobility. Its compact design combines powered and passive functionalities to ensure safe, intuitive, and adaptable operation in both home and clinical settings. 

The design of Moby was driven by key requirements identified through a needs analysis focused on enhancing independence and safety for elderly individuals in daily living. These requirements included ease of use, a lightweight design, comfort, versatility, effective sit-to-stand assistance, and the ability to maintain users in an upright posture. The subsequent sections detail how Moby's mechanical configuration and integrated features were developed to address these specific needs, ensuring safe, intuitive, and adaptable operation in both home and clinical settings.
\begin{figure*}[!th]
    \centering
    \includegraphics[width=1.0\linewidth]{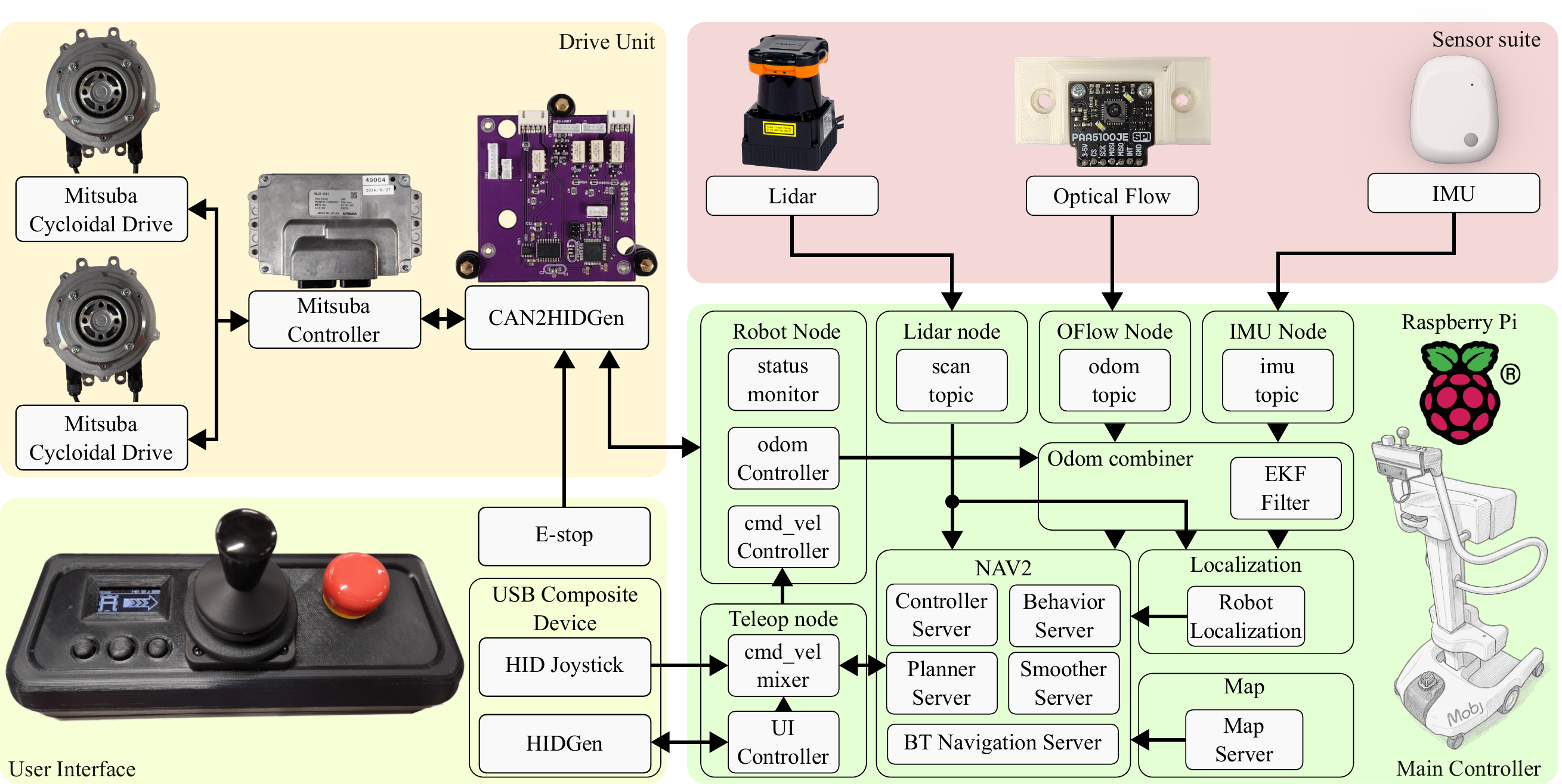}
    \caption{Overview of Moby's Control System Architecture. This diagram illustrates the interconnections between the Drive Unit, Sensor suite, User Interface, and the Main Controller, highlighting the integration of ROS nodes and navigation components.}
    \label{fig:system-design}
\end{figure*}
At its core, Moby features a mobile base equipped with two Mitsuba LA03 cycloidal drives (Gunma, Japan), each delivering 86 W of power and integrated with rheostatic brakes. These motors, combined with a four-caster configuration, create a differential drive system that allows precise maneuvering in constrained indoor environments. Importantly, the motors are back-drivable, enabling the robot to be pushed manually when not powered—offering hybrid operation as both a powered and passive walker depending on user needs and energy levels.

For postural support, a centrally mounted vertical column helps the user maintain an upright position. This column is equipped with two independently adjustable cushions—one for shin support and another for abdominal stabilization. These elements are designed to distribute pressure evenly and improve comfort, particularly during standing transitions. The shin cushion also reduces excessive knee flexion and provides additional stability when the user leans backward. Specifically, when the user leans backward, the shin cushion acts as a brace against the lower leg, preventing excessive knee flexion and transferring some of the backward forces to the robot's stable base, thereby enhancing overall stability during standing or transfers.


A key feature of Moby’s design is its ergonomic, freeform handlebar. Its freeform shape was iteratively designed based on ergonomic principles and user feedback to accommodate a variety of grip styles and hand placements, ensuring comfort and stability across different movement tasks, from initiating sit-to-stand transitions to stabilizing during short-distance walking. The handlebar allows for multiple grip styles suited to different movement tasks, such as stabilizing while walking, initiating sit-to-stand transitions, or maintaining support while navigating narrow spaces. This flexibility enables users to engage naturally with the robot based on their posture and strength.

\begin{figure}
    \centering
    \includegraphics[width=0.9\linewidth]{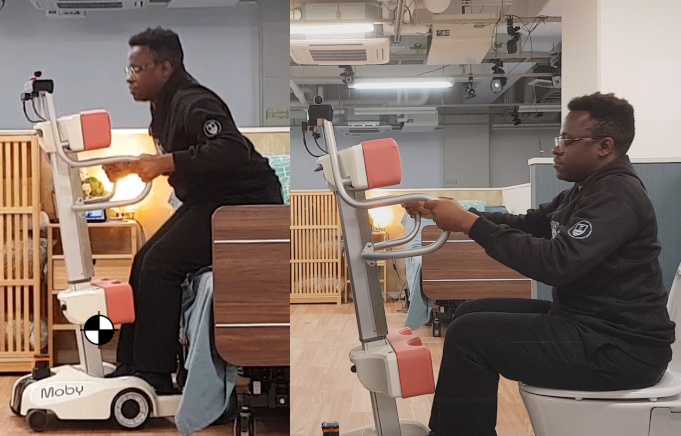}
    \caption{Typical use case of Moby when transitioning from sitting to standing, additionally, the center of gravity of the robot is shown to better explain why the user can extend past the base footprint.}
    \label{fig:figure-sit-to-stand}
\end{figure}

Moby’s ability to assist with sit-to-stand transitions is a central design goal. Its low center of gravity allows users to shift their weight forward and beyond the robot’s base without compromising balance. This functionality is especially useful for daily activities like toilet support and transfers, where stability and range of motion are crucial.  Figure~\ref{fig:figure-sit-to-stand} shows a typical transition from bed to the Moby and from Moby to the toilet by a user. The design of the Moby allows the user to never step out of the robot's platform for activities such as Toileting. The lower back shape of the Moby is also designed with this consideration, such that it can fit seamlessly to the toilet with the person's natural sitting posture. Specifically, the lower rear section of Moby's frame features a recessed or cutout design. This allows the robot to closely approach a standard toilet or bed, minimizing the gap between the user and the support surface and enabling a more natural and direct sit-to-stand or stand-to-sit transfer without requiring the user to step off the robot's platform.

Despite its robust structure, Moby remains lightweight at just 27.8 kg, allowing for easy repositioning by users or caregivers. This lightweight frame, combined with its compact footprint, enhances maneuverability in tight spaces commonly found in home or care settings. The design of Moby carefully balances the seemingly contradictory requirements of a low center of gravity for stability and a lightweight frame for maneuverability. A low center of gravity is achieved by strategically placing heavy components, such as the drive units and battery, close to the base. Simultaneously, the overall structure utilizes lightweight yet robust materials to keep the total weight lower, enabling easy repositioning by users or caregivers and enhancing maneuverability in tight spaces.

To accommodate a range of user body types and preferences, Moby’s column includes two independent adjustment mechanisms. The first allows partial collapse of the upper column to adjust the height of the control panel and upper cushion. The second allows fine-tuned adjustment of the shin support via a detent-based system, enabling proper alignment with the user’s knee height. These customizations ensure that Moby can be comfortably configured for individuals of different sizes and mobility needs.


\section{Control Systems}
The control system of the Moby robot is designed to support reliable operation in both user-driven and autonomous modes, offering a balance between intuitive human interaction and structured software architecture. It consists of modular embedded components orchestrated through custom firmware and standard ROS (Robot Operating System) robotic middleware. \autoref{fig:system-design} shows the primary components of the Moby assistive robot. 

\subsection{\underline{Operational Modes and User Interface}}

Moby operates in two principal modes: \textit{Manual Mode} and \textit{Automatic Mode}.

\begin{itemize}
    \item \textit{Manual Mode} enables direct user control via a two-axis joystick. An electronic speed selection interface allows dynamic tuning of motion sensitivity, facilitating precise handling in constrained environments. Joystick input is filtered through an acceleration and braking control module, which computes the difference between desired and current velocities to apply smooth deceleration profiles—enhancing maneuverability and safety during user-initiated motion.

    \item \textit{Automatic Mode} is facilitated via the Robot Operating System (ROS), where the robot subscribes to standard \texttt{cmd\_vel} topics. This enables full integration with ROS-based navigation stacks such as NAV2, allowing high-level trajectory planning, localization, and autonomous behavior within indoor environments.
\end{itemize}

The user interface comprises a joystick, an emergency stop (E-stop) mechanism, and a display panel equipped with three function keys. The screen provides real-time system feedback including speed levels, battery voltage, and current operating mode, supporting both user awareness and operational transparency.

\subsection{\underline{Embedded Architecture}}

The main control pipeline is implemented on a Raspberry Pi 5 (Raspberry Pi Ltd., UK), which is responsible for high-level state management and ROS integration. It handles communication with peripherals via USB HID protocols and serves as the central hub for the Robot Operating System (ROS 2) software stack. The robot publishes and subscribes to several standard ROS topics for seamless command routing and sensor data acquisition. These include  \texttt{/cmd\_vel} for velocity commands in both manual and autonomous modes, \texttt{/scan} for LiDAR sensor data, \texttt{/odom} for odometry information from motor encoders, \texttt{/tf} for coordinate frame transformations, \texttt{/battery\_state} for power levels and diagnostics, and \texttt{/robot\_description} which serves the URDF model for visualization and simulation (\autoref{fig:system-design}).

Real-time actuation tasks are offloaded to a dedicated {ATmega32U4} microcontroller, which manages low-latency control of the drive system over a Controller Area Network (CAN) bus. The CAN stack is implemented using an {MCP2515} controller and {TJA1050} transceiver. This configuration ensures deterministic communication for velocity commands, encoder feedback, and thermal/load diagnostics.

A custom-designed HID-based USB interface handles user input and screen communication with minimal latency. Additionally, {dual 24V contactors} are used for regulated battery charging and high-current load switching, with relays incorporated to provide galvanic isolation and fail-safe control.

\subsection{\underline{Motor Control and Firmware}}

The firmware architecture is distributed across the Raspberry Pi and the ATmega32U4 microcontroller. The Raspberry Pi is responsible for high-level state management and ROS integration, while the ATmega32U4 performs low-level motor control, issuing RPM commands and processing encoder data to compute odometry. Motor telemetry, including real-time temperature and load factor (expressed as a duty cycle between 0 and 100\%), is continuously monitored to prevent overheating or overcurrent conditions.

This division of labor enables stable system operation under both user-driven and autonomous control regimes without imposing real-time constraints on the main processor. During the passive sit-to-stand transitions, user can utilize their body strength for transition while the control system manages low-latency control of the drive system to provide real-time physical support and stabilization through active braking. This involves precise control of the Mitsuba Cycloidal Drives to ensure the robot's base remains stable and provides a reliable fulcrum for the user's weight shift, working in conjunction with the adjustable shin and abdominal cushions to distribute pressure and maintain balance throughout the movement.

\subsection{\underline{Sensor Integration}}

For obstacle detection and spatial awareness, the robot incorporates a {Hokuyo UTM-X001s LiDAR} (Osaka, Japan), mounted above the central control PCB. This sensor provides 2D range data and integrates directly with ROS to support navigation, mapping, and safety-aware trajectory control. 

\begin{figure*}[t!]
    \centering
    \begin{subfigure}[t]{0.45\textwidth}
        \centering
        \includegraphics[width=1.0\linewidth]{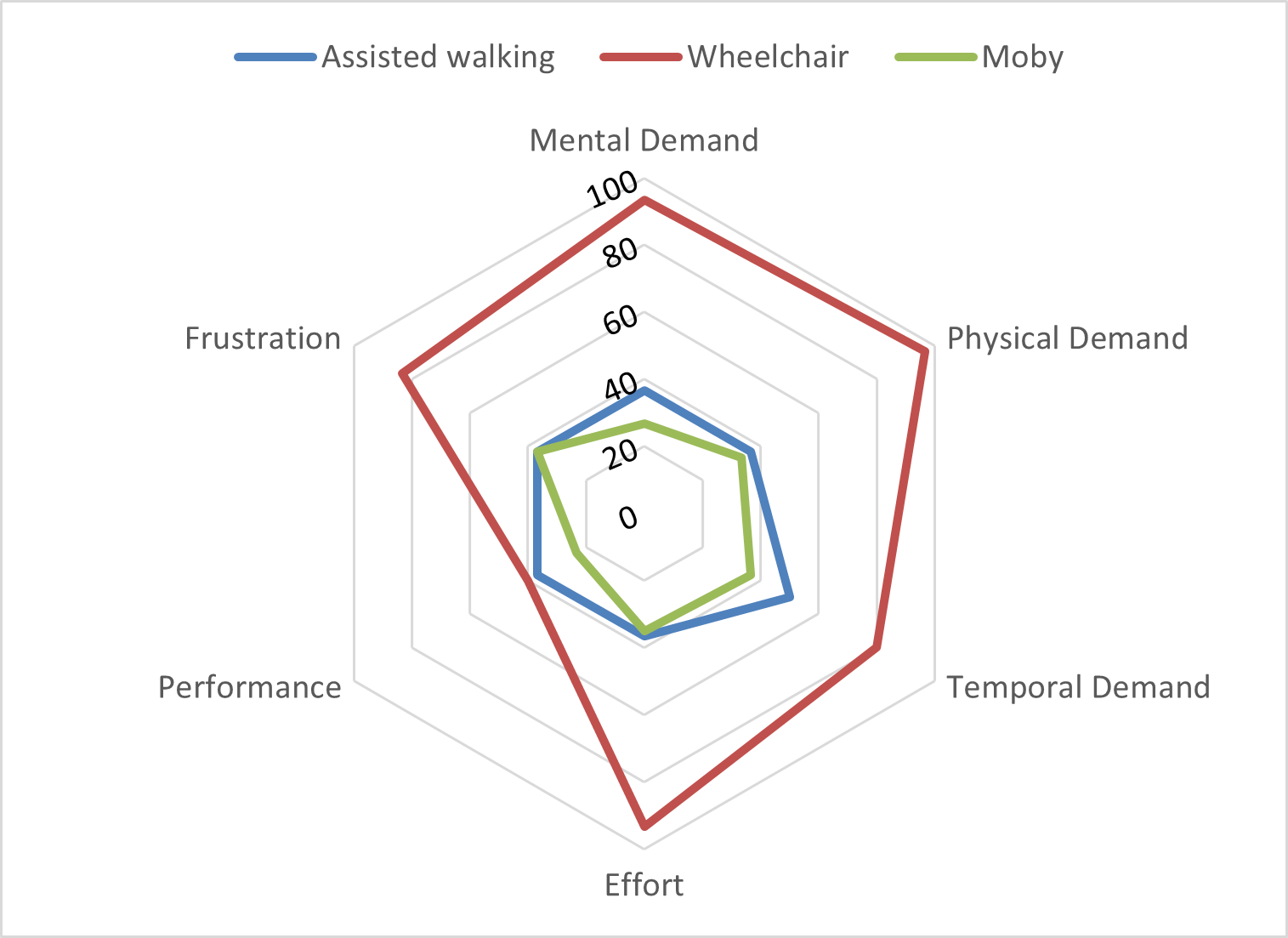}
        \caption{NASA-TLX Results}
        \label{fig:figure-resultstlx}
    \end{subfigure}%
    ~
    \begin{subfigure}[t]{0.45\textwidth}
        \centering
        \includegraphics[width=1.0\linewidth]{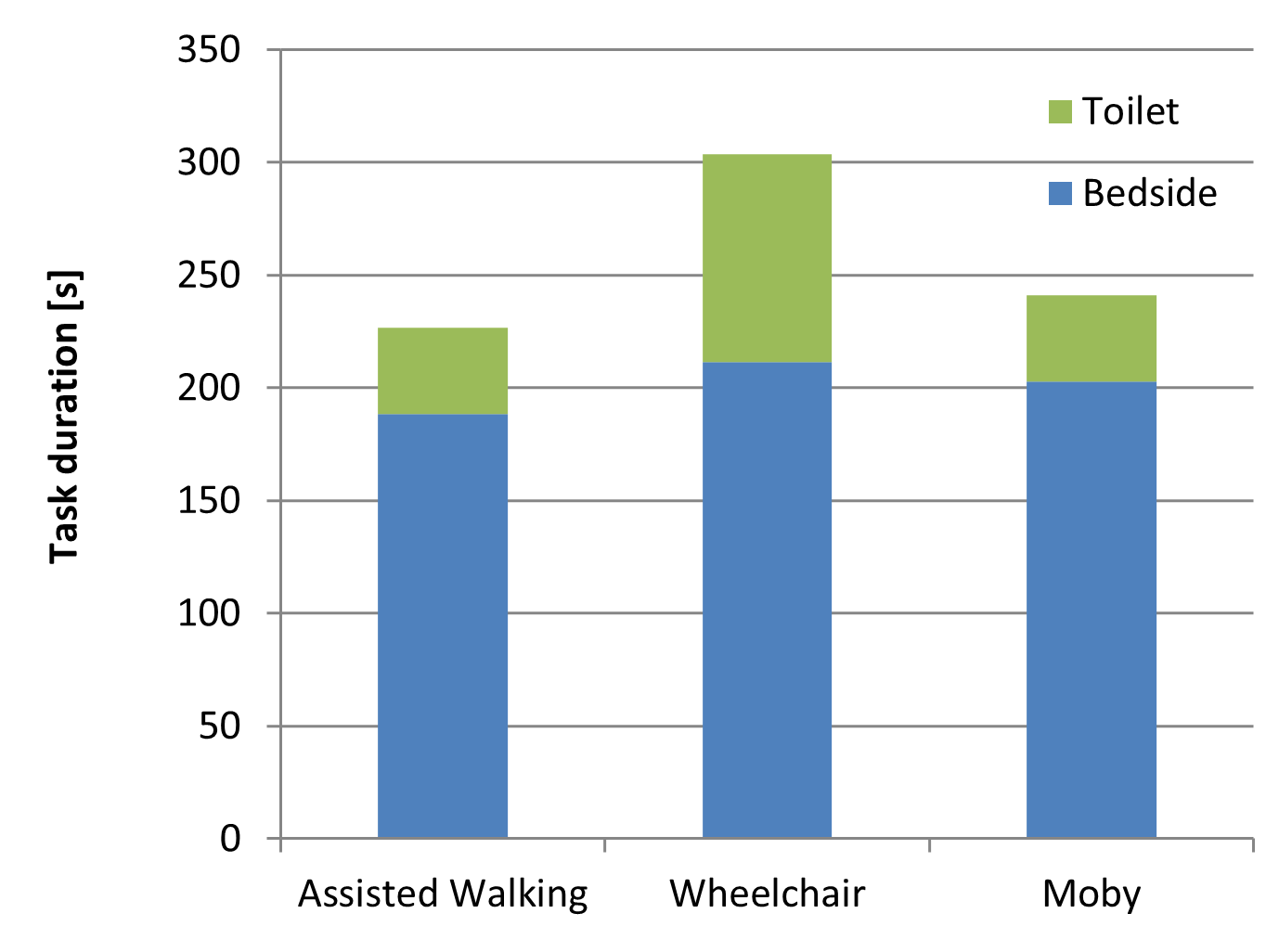}
        \caption{Comparison of time used for toileting with different assistive methods}
        \label{fig:figure-timeresults}
    \end{subfigure}
    \caption{Experimental results of performance evaluation of Moby against Wheelchair use and Assisted walking}
    \label{fig:experimental-results}
\end{figure*}

\section{Evaluation and use case scenarios}
Enabling safe, independent execution of daily hygiene routines is one of the most impactful and challenging goals of assistive robotics. In particular, toileting routines often require multiple transitions—typically from bed to wheelchair, then from wheelchair to toilet—which introduce physical strain, cognitive load, and a heightened risk of falls. In conventional approaches, such as assisted walking, a caregiver is required to provide physical support during transitions, especially while sitting down or standing up from the toilet. Moby was designed to eliminate this dependency by converting complex sit-to-sit transitions into simplified, mechanically supported stand-to-sit movements, allowing users to perform these tasks autonomously.

\subsection{\underline{Experimental Evaluation}}

To evaluate Moby’s effectiveness, we conducted a comparative study of three mobility strategies:

\begin{enumerate}
    \item Assisted Walking (with caregiver)
    \item Conventional Wheelchair Use
    \item Moby-Assisted Mobility
\end{enumerate}

Each of three participants performed a full toilet transfer routine under all three conditions. The participants were healthy individuals and would move from a bedroom with the bed to the toilet that passes through a small corridor environment. The total distance to the toilet is approximately 20m. The evaluation used both \textit{objective metrics} (task time) and \textit{subjective workload feedback} (NASA-TLX). All experimental procedures involving human participants were conducted in accordance with the ethical guidelines. All participants provided their informed consent prior to participation.

\subsubsection{\underline{Completion Time}}
The total task duration was measured for a complete toilet transfer routine, which encompassed the entire process from the participant starting in bed, moving to the toilet (approximately 20m distance, including navigation through a small corridor environment), performing the toilet transfer, and returning to the bedside. Measurement for each trial began from the moment the participant initiated the transfer from the bed (i.e., attempting to stand up or engage with the assistive device) and concluded once they were seated back on the bed. In the Moby and Wheelchair conditions, the respective devices were initially positioned near the bed, ready for immediate use. The 'Bedside Transfer' time accounts for the initial transfer from the bed and subsequent maneuvering to exit the bedroom, while 'Toilet' time specifically covers the maneuver to approach, transfer to, and depart from the toilet, including the final return to the bed.
As shown in Fig.~\ref{fig:figure-timeresults}, the total task durations were:
\begin{itemize}
    \item Wheelchair: \textbf{~304 s}
    \item Moby: \textbf{~241 s}
    \item Walking: \textbf{~226 s}
\end{itemize}

While assisted walking showed the shortest time, it required \textit{constant caregiver involvement}, particularly during toilet transitions. Moby, on the other hand, enabled \textit{completely independent execution} of the same routine, with only a 6.6\% increase in total time. On the other hand Moby showed  a 50\% reduction in time as compared to wheelchair case. This suggests that Moby is particularly effective in streamlining toilet-related transfers, likely due to its standing support design, which eliminates the need for complex repositioning.

Breaking down the transitions:
\begin{itemize}
    \item \textbf{Toilet Transfer}: Moby and walking both required \textbf{~38 s}, while wheelchair took \textbf{~92 s}.
    \item \textbf{Bedside Transfer}: All three methods showed comparable times (~200 s).
\end{itemize}

\noindent\textbf{Insight:} Moby achieves time efficiency equivalent to human-assisted walking, while removing caregiver dependence and reducing transition complexity.

\subsubsection{\underline{Technical Interpretation}}

Moby’s performance advantage arises from its \textit{ergonomically designed standing support structure}, which stabilizes the user’s \textit{center of gravity} and distributes load across \textit{shin and abdominal contact points}. While direct biomechanical measurements were not conducted in this study, the significant reduction observed in the `Physical Demand' and `Effort' dimensions of the NASA-TLX results  serves as an indirect indicator supporting this assertion. Users reported a subjectively lower physical workload when utilizing Moby, suggesting that the robot's design effectively mitigates some of the biomechanical demands associated with transfers.

During the experiment, Moby was manually operated by the user using the joystick interface. The observed performance therefore reflects the ease and responsiveness of the control system under user-driven input. The robot’s ergonomic posture and low-friction drive base allowed for intuitive maneuvering, even during precise tasks such as bedside alignment and toilet positioning. This highlights Moby’s suitability for semi-active use, where minimal training or coordination is required for safe operation.

\subsubsection{\underline{NASA-TLX Results}}

As summarized in Fig.~\ref{fig:figure-resultstlx}, Moby outperformed both alternatives across all six NASA-TLX dimensions:

\begin{itemize}
    \item \textbf{Mental Demand:} Moby (27) $<$ Walking (37) $<$ Wheelchair (93)
    \item \textbf{Physical Demand:} Moby (33) $<$ Walking (37) $<$ Wheelchair (97)
    \item \textbf{Temporal Demand:} Moby (37) $<$ Walking (50) $<$ Wheelchair (80)
    \item \textbf{Effort:} Moby (35)$<$ Walking (37) $<$ Wheelchair (93)
    \item \textbf{Performance (lower is better):} Moby (23) $<$ Walking (37) $<$ Wheelchair (40)
    \item \textbf{Frustration:} Moby $=$ Walking (37) $<$ Wheelchair (83)
\end{itemize}

Moby reduces both cognitive and physical workload while improving perceived autonomy and confidence. The reduction in \textit{mental demand} is particularly significant, highlighting that users felt safe and in control during operation, without the need for continuous monitoring or support. While these results are based on a small participant group and formal statistical significance testing was not performed, the observed reduction in mental demand across all participants strongly suggests an improved subjective sense of safety and control during Moby's operation compared to the other methods. Future studies with larger cohorts will enable more robust statistical analysis to confirm these findings.

\subsection{\underline{IoT Integration and Autonomous Deployment}}

In a deployment demonstration at the Aobayama Living Lab (Tohoku University), Moby was integrated with {Amazon Alexa}, ROS 2-based {NAV2}, and smart home infrastructure to support {autonomous nighttime toilet routines}\footnote{\url{bit.ly/moby-robot}}. 

A user can initiate the sequence via voice command. Moby autonomously navigates from its docking station to the bedside, where the user boards independently. The robot then transports the user to the toilet without requiring manual control or dismounting. After the task, Moby returns the user to bed and re-docks autonomously. During the whole task the user is on the Moby (including the toilet task) and this demonstrates significant ease in doing assisted daily living activities independently by the user. 

This use case demonstrates Moby’s potential as a \textit{fully integrated assistive system}, combining robotics, autonomy, and voice interaction to provide round-the-clock support for high-risk tasks—independently of caregiver presence.

\section{Discusssions}

This study demonstrates that Moby not only reduces physical and cognitive effort during critical hygiene tasks but does so while enabling {full user autonomy}—a clear improvement over both wheelchair and caregiver-dependent methods. Its technical design enables smoother, safer transfers; its software stack provides reliable autonomous operation; and its ergonomic posture enhances user confidence and comfort.

As robotic mobility aids become more integrated into home environments, systems like Moby represent a promising direction for empowering elderly individuals with dignity, safety, and independence in their daily lives.

Despite these promising outcomes, several limitations remain. The current implementation relies on joystick-based manual control, which may be challenging for users with limited dexterity or cognitive load tolerance. Future versions will benefit from adaptive control strategies such as voice commands, shared autonomy, or intent recognition to broaden accessibility. One limitation of our experimental evaluation was the absence of a direct comparison with conventional walkers. While Moby shares the goal of supporting individuals with some standing and walking ability, its key advantages, such as active sit-to-stand assistance, adaptable postural support, and autonomous navigation capabilities, distinguish it from passive walkers. Future work will explore comparative studies with a broader range of assistive devices, including various types of walkers, to more fully quantify these distinctions.

Additionally, the experiments were conducted with a small participant group in a controlled setting. Longer-term studies in real home or care environments are necessary to assess robustness, user variability, and system reliability under daily use conditions.

While Moby demonstrated autonomous operation in a smart home deployment, this requires a mapped environment and consistent infrastructure. Broader deployment will require enhanced perception, localization, and environment generalization capabilities.

The psychological and social benefits of maintaining a standing posture—such as improved eye-level communication and engagement in routine tasks—are promising, but warrant deeper study through longitudinal trials.

Finally, future developments will focus on integrating natural language interfaces and multi-modal scene understanding to position Moby as not only a mobility aid, but a conversational assistant capable of supporting routines, memory, and contextual guidance.

In summary, Moby shows strong potential as a foundation for next-generation assistive robots, with further work required to transition from research platform to daily companion in real-world settings.

\section{CONCLUSIONS}
This work introduced \textit{Moby}, a standing support robot designed to empower elderly individuals with greater autonomy in daily life. Unlike traditional seated mobility aids such as wheelchairs—which often symbolize physical decline and dependency—Moby enables users to remain upright, bridging the gap between full independence and total reliance on external support. In doing so, it establishes a new class of assistive mobility: one that preserves dignity, promotes physical engagement, and enhances social participation.

Through experimental evaluation and scenario-based demonstrations, we showed that Moby significantly reduces the physical and cognitive demands of high-risk tasks such as toilet transfers. Importantly, it allows users to perform these routines independently—without caregiver involvement—while maintaining time efficiency and user confidence. Its ergonomic support posture, intuitive control system, and adaptability make it a practical and user-friendly solution for mobility-challenged individuals.

More than a mobility device, Moby reinforces {self-efficacy}—the belief in one's ability to manage tasks without assistance. This fosters not just functional independence, but also emotional well-being. By supporting users in a standing posture, Moby facilitates more natural social interactions at eye level, improving communication and reducing social detachment. It also enables participation in light household activities, interaction with the environment, and movement through shared spaces—turning assistive mobility into a tool for active living.

Moby is currently a research platform, and ongoing development will focus on {long-term deployment and studies in care environments}, including senior homes, personal residences, and hospitals. These studies will help uncover the robot's full potential—as well as its limitations—under real-world conditions, where users’ physical and emotional needs evolve over time.

Looking ahead, Moby will incorporate {natural language-based control and interaction}, enabling it to act as a conversational assistant. By leveraging large language models and multi-modal sensor input, Moby could help users remember schedules, answer queries, or even provide context-aware information based on its navigation system, camera feed, and environmental sensors. This evolution will position Moby not just as a mobility aid, but as a trusted, communicative partner in daily living.

As global aging accelerates and caregiver shortages intensify, systems like Moby may reshape the future of assistive robotics—redefining aging as a phase of continued autonomy, dignity, and meaningful human connection.



\addtolength{\textheight}{-12cm}   




\section*{ACKNOWLEDGMENT}
We would like to thank Paramount Bed Co. Ltd., Japan for assisting in the research and development of Moby. 
This work was partially supported by JST Moonshot R\&D [Grant Number JPMJMS2034] and JSPS Kakenhi [Grant Number JP24K07399].


\bibliographystyle{ieeetr}
\bibliography{root}

\end{document}